\documentclass[10pt,twocolumn,letterpaper]{article}

\usepackage{cvpr}
\usepackage{times}
\usepackage{epsfig}
\usepackage{graphicx}
\usepackage{amsmath}
\usepackage{amssymb}
\usepackage{verbatim}
\usepackage{subfigure}
\usepackage{rotating}
\usepackage{color}

\usepackage{multirow}

\cvprfinalcopy 

\def\cvprPaperID{337} 

\ifcvprfinal\fi
\begin{document}


\title{Shadow Estimation Method for \\``The Episolar Constraint: Monocular Shape from Shadow Correspondence''}

\author{
Austin Abrams\textsuperscript{1}, Chris Hawley\textsuperscript{1}, Kylia Miskell\textsuperscript{1},
Adina Stoica\textsuperscript{1}, Nathan Jacobs\textsuperscript{2}, Robert Pless\textsuperscript{1} \\
\begin{tabular}{c@{\hspace{.5in}}c}
  \textsuperscript{1}Washington University in St Louis~~~~~~\textsuperscript{2}University of Kentucky  \\
  \end{tabular}
}

\maketitle

\begin{abstract}
Recovering shadows is an important step for many vision algorithms.  
Current approaches that work with time-lapse sequences are limited to simple thresholding heuristics.
We show these approaches only work with very careful tuning of parameters, and do not work well for
long-term time-lapse sequences taken over the span of many months.  
We introduce a parameter-free expectation maximization approach which simultaneously estimates shadows, albedo, 
surface normals, and skylight.  
This approach is more accurate than previous methods, works over both very short and very long sequences, and is robust
to the effects of nonlinear camera response.
Finally, we demonstrate that the shadow masks derived through this
algorithm substantially improve the performance of sun-based
photometric stereo compared to earlier shadow mask estimation.
\end{abstract}

\section{Introduction}

\begin{figure}[t]
\begin{center}

\subfigure[ Example image from a time-lapse]{
											\includegraphics[width=0.22\textwidth]{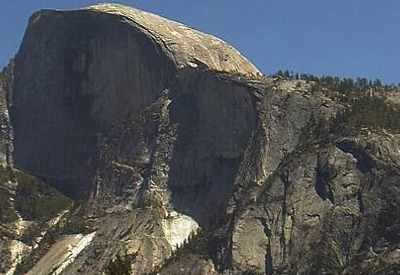} 	\label{frontpage:example}}
\subfigure[Algorithm from \cite{abrams2012helio}]{
											\includegraphics[width=0.22\textwidth]{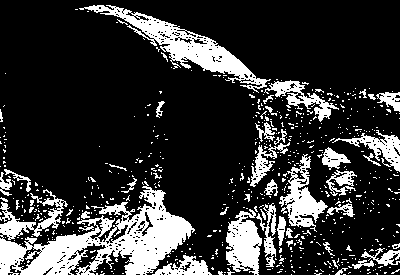} \label{frontpage:helio}}								
\subfigure[Algorithm from \cite{sunkavalli2007timelapse}]{
											\includegraphics[width=0.22\textwidth]{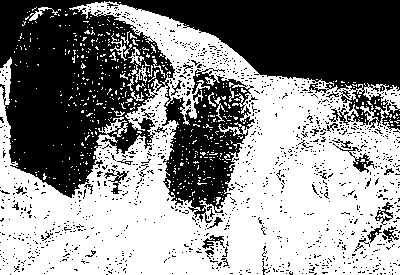} \label{frontpage:ftlv}}		
\subfigure[Our approach]{
						 \includegraphics[width=0.22\textwidth]{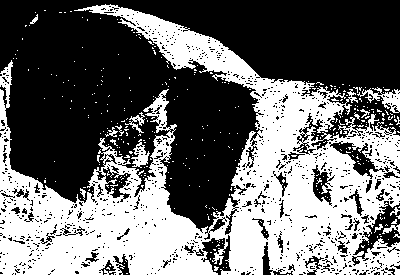} \label{frontpage:em}}

\caption{Given a set of time-lapse imagery~\subref{frontpage:example}, we wish to classify each pixel at each time as being under shadow or not.  
Previous work (\subref{frontpage:helio} and \subref{frontpage:ftlv}) demonstrates some success in simple thresholding
heuristics, but these fail for oblique lighting directions and poor selection of tuning parameters.  Our
approach  \subref{frontpage:em} is parameter-free, robust to changing lighting conditions, and outperforms previous methods.}
\label{fig:frontpage}
\end{center}
\end{figure}

Shadows are a critical component of image formation.  They are one of
the largest causes of appearance change in outdoor scenes.  Across
many problem domains, invariance to lighting drives
choices in image pre-processing and feature selection.

Recently, several works have aimed to understand how to model outdoor image formation through time~\cite{abrams2012helio, ackermann2012ps, 
sunkavalli2007timelapse}, all of which explicitly model shadows.  In each of these works, the authors use some variant of 
heuristic thresholding to estimate the
\emph{shadow-or-not classification problem}: given many images of a static scene under varying illumination,
which pixels are directly illuminated at which times?   

Previous methods are heuristics that require parameter tuning and
still often fail for data drawn from weeks/months instead of just a
single day.  In this work we focus on the problem of solving for the
shadow-or-not classification problem in real scenes as captured by outdoor webcams, over a variety of lighting directions and timespans, without parameter tweaking.

We use an expectation-maximization approach which estimates the
expected intensity of an image under direct sunlight and under shadow.  Our approach explicitly models the sun as a moving light
source and recovers the Lambertian world most consistent with that lighting.  We find that this model more aptly captures the real world
than simple thresholding, works well for very short and very long sequences, and robustly handles a variety of real-world distortions such
as nonlinear camera response.

We offer three novel contributions.  First, the introduction of a parameter-free shadow estimation
procedure for time-lapse sequences of outdoor scenes over both very long and short time frames.  Second, we characterize
how our algorithm performs over varying lengths of time and under the effect of nonlinear radiometric response, 
details which are not modeled in our formulation in order to make the computation more tractable.  Finally, we
explore how our algorithm performs on real-world cameras and introduce the Labeled Shadows in the Wild dataset (7 
scenes consisting of 50 images per scene with ground truth shadow labels) to quantitatively compare 
shadow classification tasks in real outdoor settings for future studies.  Our code and this data set will be publicly shared.

\section{Related Work}

Our method can be seen as an approach for simultaneous shadow estimation and photometric stereo~\cite{woodham1980photometric}.  
Most active research in photometric stereo is focused on \emph{uncalibrated} data sets, where the light direction and intensity 
are unknown.  In contrast, we use webcams with known geolocation and timestamp, so we can use a solar position lookup~\cite{reda2003sunposition}
to recover the lighting direction.

A large body of work solves the photometric stereo problem in the presence of shadows by treating them as noisy measurements.
Wu et al.~\cite{wu10robust} treat
shadows as large-but-sparse errors in the Lambertian model and frame photometric stereo as a low-rank matrix factorization problem.
This approach offers robust estimation in the face of specularities, sparse shadows, and large outliers.  Wu and Tang~\cite{wu2006em}
use an expectation maximization approach to simultaneously solve the photometric stereo problem and estimate a pixel-wise weight defining if each pixel satisfies the Lambertian model.  Chandraker et al.~\cite{chandraker07shadowcuts} isolate groups of pixels that are 
simultaneously lit by a common light source through a Markov Random Field to support spatial smoothness. Sunkavalli et
al.~\cite{sunkavalli10subspace} extend this approach to the uncalibrated case.

These works 
introduce a variety of parameters: \cite{wu10robust} uses a tradeoff between a low-rank and sparse-error recovery, 
\cite{wu2006em} specifies Gaussian bandwidths and a penalization cost for breaking the Lambertian model~\footnote{The authors of \cite{wu2006em}
remark that most reasonable automated choices of this cost give approximately the same result, suggesting that no \emph{user-specified} 
parameter is truly necessary.}, \cite{chandraker07shadowcuts} uses a 
tradeoff between a data and smoothness term, and \cite{sunkavalli10subspace} uses various thresholding parameters in a RANSAC 
setting.  In pursuit of a truly automatic method, our approach contains no parameters whatsoever.

Our approach differs from all of the above works by treating shadows not as noise to be detected and ignored,
but rather by explicitly modeling the shadow process within the image formation model.  This is an important step 
for automatically-interpreting outdoor imagery, where the contribution of ambient light is substantial, and 
shadowed pixels are frequent.  This explicit modeling has the benefit of removing all parameters from the 
algorithm.

Prior experiments report results on very controlled environments, taken in a dark room with known
camera settings.  Notable exceptions include Ackermann et al.~\cite{ackermann2012ps} and
Abrams et al.~\cite{abrams2012helio} which solve the photometric stereo problem for outdoor webcams.  Sunkavalli et al.~\cite
{sunkavalli2007timelapse} also present an outdoor image formation model, but because they work with only a single day of
imagery, they only recover a 1-D projection of each pixel's surface normal.  

\subsection{Shadow Estimation in Time-Lapse Sequences}

Most previous approaches design a threshold for each pixel by analyzing that pixel's intensity trajectory through time.
This section describes our notation, the prior approaches, and their parameters.

We denote an image taken at time $t$ as a $p$-element vector $I_t$ between 0 and 255, where $p$ is the number of non-sky pixels.
The goal is to take a set of $n$ images $I_1, \dots, I_n$ and recover a sunny-or-not binary classification for 
each image, $S_1, \dots, S_n$.  We index a pixel $\mathbf{x}$ at time $t$ as
$I_t(\mathbf{x})$ or $S_t(\mathbf{x})$.  By convention we denote that if some pixel $\mathbf{x}$ is
directly lit at time $t$, then $S_t(\mathbf{x}) = 1$, otherwise $S_t(\mathbf{x}) = 0$.  

\paragraph{Factored Time-Lapse Video}
The approach in \cite{sunkavalli2007timelapse} observes that
over the span of a day, most pixels are under shadow at least 20\% of the time.  This leads
them to a heuristic that finds the median of the shadowed intensity (the
10th percentile pixel), then choosing a threshold at 1.5 times that
value.  This is the approach used in~\cite{ackermann2012ps} for outdoor
photometric stereo.  We later explore how various settings of these
parameters affect results on the shadow-or-not classification task, so we generalize their approach to handle arbitrary scalar multipliers $\theta_k$ and
percentiles $\theta_p$:

\begin{equation}
S_t(\mathbf{x}) \leftarrow \begin{cases} 1 & I_t(\mathbf{x}) \geq \theta_k \operatorname{per}(I(\mathbf{x}), \theta_p)\\
 									     0 & \text{otherwise} \end{cases}
\end{equation}
where $\operatorname{per}(A, \theta_p)$ returns the of bottom $\theta_p$th percentile value of the set $A$ of grayscale intensities, and
$I(\mathbf{x})$ is shorthand to denote $\{I_t(\mathbf{x})\}_{t=1}^n$.  

\paragraph{Heliometric Stereo}
The approach presented in \cite{abrams2012helio} also works by simple thresholding, but allows the threshold to adaptively change
from frame to frame (assuming the images are listed in chronological order).  This adaptive approach
attempts to model the changing light intensity across seasons; this is important for long-term time-lapses 
because a shadowed pixel in the summer---where the sun is highest in the sky---might be brighter than a 
lit pixel in the winter, so often a single threshold does not work.

For each pixel $\mathbf{x}$, their approach defines two centroids: the expected intensity of that pixel when it is directly
lit and under shadow, denoted $E_L$ and $E_S$ respectively.
The centroids $E_L$ and $E_S$ are initially set by taking the top and bottom $\theta_p$ percentiles of the
image sequence.  For each image from $t = 1 \to n$, if the difference from $E_L(\mathbf{x},t-1)$ to $I_t(\mathbf{x})$
is smaller than the difference from $E_S(\mathbf{x},t-1)$ to $I_t(\mathbf{x})$, then update 
\begin{equation}
E_L(\mathbf{x},t) \leftarrow E_L(\mathbf{x}, t-1) \theta_\lambda + I_t(\mathbf{x}) (1-\theta_\lambda).
\end{equation}
Otherwise, update 
\begin{equation}
E_S(\mathbf{x},t) \leftarrow E_S(\mathbf{x}, t-1) \theta_\lambda + I_t(\mathbf{x}) (1-\theta_\lambda),
\end{equation}
where $\theta_\lambda \in [0,1]$ is a parameter that defines how quickly these centroids can change.
Finally, this centroid-updating step is reversed, from $t = n \to 1$ to lessen the effect of initialization
on centroids close to $t = 1$.  The final shadow labeling is determined by whether the original image fits the
expectation of a shaded or directly-lit pixel:

\begin{equation}
S_t(\mathbf{x}) \leftarrow \begin{cases} 1 & |I_t(\mathbf{x}) - E_L(\mathbf{x},t)| \leq |I_t(\mathbf{x}) - E_S(\mathbf{x},t)| \\
										 0 & \text{otherwise} \end{cases}
\end{equation}

In summary, at the span of one or a few days, the approach in~\cite{sunkavalli2007timelapse} performs well, and
over the span of a few months, the more complicated heuristic in \cite{abrams2012helio}
does somewhat better at capturing shadows over long time periods, but
at the cost of an additional parameter $\theta_\lambda$.  We show in Section~\ref{sec:results} that
more formal modeling of the image formation process gives improved
results over even the optimal parameter settings.

\section{Parameter-Free Shadow Estimation}
\label{sec:methods}

In all cases, previous shadow estimation procedures do not attempt to model changing lighting direction.  We argue that this is an unnecessary
restriction, since sun position algorithms~\cite{reda2003sunposition} very accurately estimate the sun position 
given its capture time and geolocation, which itself can be determined automatically by a variety of
methods~\cite{jacobs11geolocate,junejo2010geolocate, wu2010geolocate}.

Therefore, we assume that a camera has been geolocated and accurately timestamped to recover per-image lighting
directions $L_1, \dots, L_t, \dots, L_n$ as three-dimensional unit vectors.  Given this information, we develop
an expectation maximization approach which solves for the shadows most consistent with a Lambertian assumption.
Borrowing from the image formation models of~\cite{abrams2012helio, ackermann2012ps}, we use a
simple Lambertian model to represent our scene:
\begin{equation}
I_t(\mathbf{x}) \approx \rho(\mathbf{x}) (\max(L_t \cdot N(\mathbf{x}),0) S_t(\mathbf{x})  + A(\mathbf{x}))
\label{eqn:lambertian_model}
\end{equation}
where $\rho(\mathbf{x}), N(\mathbf{x}),$ and $A(\mathbf{x})$ are the albedo, normal vector, and skylight (ambient light
contribution) of a pixel $\mathbf{x}$, respectively.  To handle color, we represent $\rho
(\mathbf{x})$ as an RGB vector, while the skylight remains grayscale.  Therefore, our goal is to estimate the 
unknown albedo, surface normal, ambient light, and a shadow labeling for each pixel, given the original imagery and associated
lighting directions.

Compared to~\cite{abrams2012helio, ackermann2012ps}, we do not include any time-varying unknowns (such as light intensity, exposure, or ambient intensity), the 
camera's unknown radiometric response, or attempt to handle 
non-Lambertian surfaces.  Although it would be possible to extend this model to handle these unknowns, using a simpler model
results in a simpler algorithm which already outperforms current state-of-the-art shadow estimation approaches.
This simpler model is efficient enough to be used as pre-processing
for optimization over a more complete image formation model.

Our approach alternates between fitting the per-pixel parameters
$\rho(\mathbf{x}), N(\mathbf{x}),$ and $A(\mathbf{x})$, and updating the shadow volume $S_t(\mathbf{x})$.

\subsection{Expectation Step}
In the expectation step, we aim to find the expected albedo, normals, and skylight, given the shadow volume.  
Performing this expectation over RGB images gives a nonlinear problem, because the normal vector must be
constrained to unit length.
In the case of grayscale images, a change in notation expresses the Lambertian model as an independent system of linear equations
for each pixel $\mathbf{x}$.
Writing $\hat \rho$ as the grayscale albedo, $[a(\mathbf{x}),b(\mathbf{x}),c(\mathbf{x})]^\top = \hat \rho(\mathbf{x}) N(\mathbf{x})$ and $d(\mathbf{x}) = \hat \rho(\mathbf{x}) A(\mathbf{x})$, we solve the following
linear system:
\begin{equation}
\left [
\begin{array}{ccc}
S_1(\mathbf{x}) L_1^\top & & 1\\
&\vdots & \\
S_n(\mathbf{x}) L_n^\top & & 1
\end{array}
\right ]
\left [
\begin{array}{c}
a(\mathbf{x})\\b(\mathbf{x})\\c(\mathbf{x})\\d(\mathbf{x})
\end{array}
\right ]
=
\left [
\begin{array}{c}
I_1(\mathbf{x})\\
\vdots  \\
I_n(\mathbf{x})
\end{array}
\right ].
\label{eqn:leastsquares}
\end{equation}
This $n \times 4$ system of equations therefore solves for the Lambertian model with a skylight term (i.e. surface
normal, skylight, and grayscale albedo)  most consistent with the data, for a single pixel $\mathbf{x}$.
After solving for auxiliary variables $a,b,c,$ and $d$, we recover the albedo, normal, and skylight as:
\begin{equation}
\hat \rho(\mathbf{x}) = \sqrt{a(\mathbf{x})^2 + b(\mathbf{x})^2 + c(\mathbf{x})^2},
\end{equation}
\begin{equation}
N(\mathbf{x}) = \frac{[a(\mathbf{x}), b(\mathbf{x}), c(\mathbf{x})]^\top}{\hat \rho(\mathbf{x})},
A(\mathbf{x}) = \frac{[d(\mathbf{x})]}{\hat \rho(\mathbf{x})}.
\end{equation}

To handle color images, we first run the algorithm on grayscale images to recover grayscale normals and skylight,
and solve for the best color albedo through a closed-form solution:
\begin{equation}
\rho_C(\mathbf{x}) = \frac{1}{n} \sum_{t=1}^n \frac{I^C_t(\mathbf{x})}{\max(L_t \cdot N(\mathbf{x}),0) S_t(\mathbf{x})  + A(\mathbf{x})},
\end{equation}
where $C \in \{R,G,B\}$ is the color channel of the albedo or image.  

Notice that only here do we take this opportunity to strictly
enforce non-negative Lambertian lighting.  Ideally, we would also want to solve for the surface normal in Equation~\ref
{eqn:leastsquares} that satisfies this constraint.  Further, since the hinge function $\max(x,0)$ is convex,
enforcing such a constraint would yield a globally-optimal solution.  However, solving with respect
to the hinge loss increases runtime dramatically, and in practice, the shadow volume $S_t(\mathbf{x})$ quickly stabilizes to 
cover all times when the pixel is under shadow, including attached shadows when $L_t \cdot N(\mathbf{x}) < 0$, thus zero-ing out 
the Lambertian term without use of the hinge function.

\subsection{Maximization Step}
In the maximization step, we aim to find the maximum-likelihood classification of a pixel $\mathbf{x}$ at time $t$ as being in 
shadow or not, given the current estimates of albedo, normal, and skylight.  In our case, we simply evaluate the quality of the 
reconstruction in each case, and choose the best:
\begin{eqnarray}
 \hspace{-1em}r_1 \hspace{-0.5em} &\leftarrow& \hspace{-1em}||I_t(\mathbf{x}) - \rho(\mathbf{x}) (\max(L_t \cdot N(\mathbf{x}),0) + A(\mathbf{x}))||^2\\
 \hspace{-1em}r_0 \hspace{-0.5em} &\leftarrow& \hspace{-1em}||I_t(\mathbf{x}) - \rho(\mathbf{x}) A(\mathbf{x})||^2
\end{eqnarray}
\begin{equation}
S_t(\mathbf{x}) \leftarrow \begin{cases} 1 & r_1 \leq r_0 \\ 0 & \text{otherwise} \end{cases} 
\end{equation}

\begin{figure}[t]
\begin{center}

\subfigure[Example images]{
			\includegraphics[width=0.21\textwidth]{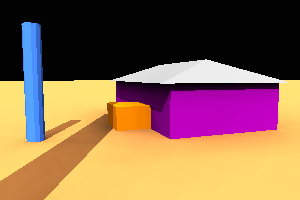}
			\includegraphics[width=0.21\textwidth]{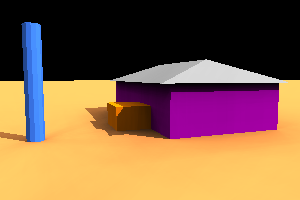}			 
			 \label{fig:synthetic:example}}
			 
\subfigure[Ground truth $S$]{\includegraphics[width=0.15\textwidth]{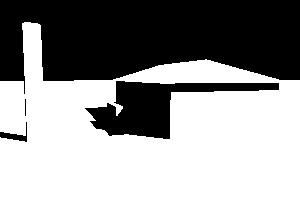}
			\label{fig:synthetic:gtshadows}}
\subfigure[Ground truth $\rho$]{\includegraphics[width=0.15\textwidth]{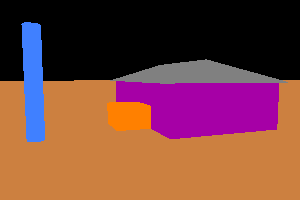}
			\label{fig:synthetic:gtalbedo}}
\subfigure[Ground truth $N$]{\includegraphics[width=0.15\textwidth]{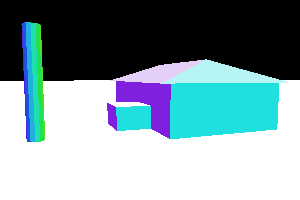}
			\label{fig:synthetic:gtnormals}}
			
\subfigure[Recovered $S$]{\includegraphics[width=0.15\textwidth]{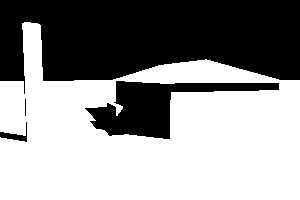}
			\label{fig:synthetic:recshadows}}
\subfigure[Recovered $\rho$]{\includegraphics[width=0.15\textwidth]{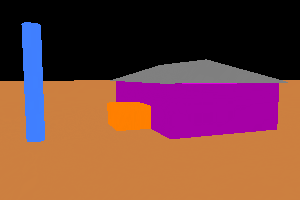}
			\label{fig:synthetic:recalbedo}}
\subfigure[Recovered $N$]{\includegraphics[width=0.15\textwidth]{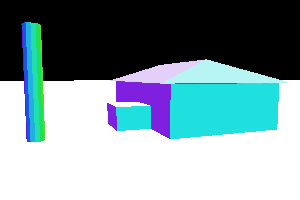}
			\label{fig:synthetic:recnormals}}
\caption{Experiments with synthetic data.  We use a rendering pipeline to create 300 images using a year's
		 worth of simulated lighting directions \subref{fig:synthetic:example}.  Our recovered results
		 from this sequence match the ground truth results almost exactly.}
		 \label{fig:synthetic_reconstruct}

\end{center}
\end{figure}
\begin{figure*}[t]
\begin{center}
\begin{tabular}{c|c}
\subfigure[]{\includegraphics[height=0.14\textheight]{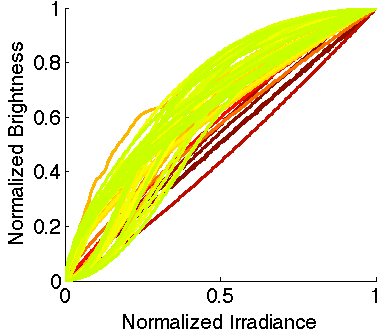}
			 \includegraphics[height=0.14\textheight]{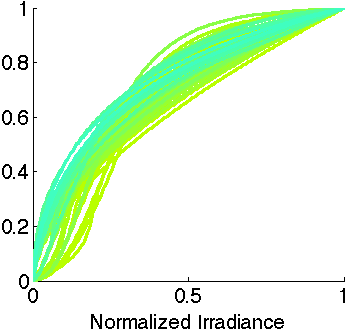} 
			 \includegraphics[height=0.14\textheight]{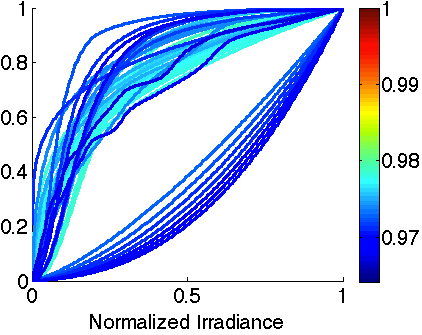}  \label{subfig:responsefns}}&
\subfigure[]{\includegraphics[height=0.14\textheight]{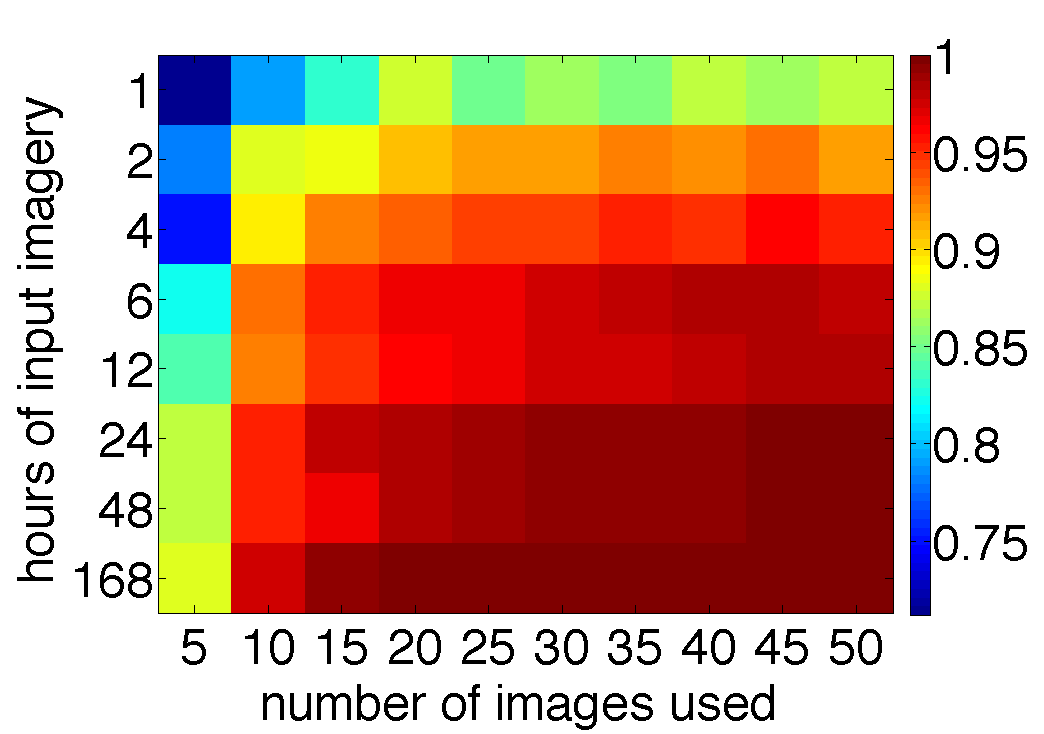} \label{subfig:datasize}}
\end{tabular}
\caption{Sensitivity experiments of the proposed approach on various synthetic datasets. In~\subref{subfig:responsefns}, we 	
		 use 300 images from a simulated year-long period and distort the input sequence by a variety of nonlinear camera response 
		 functions from~\cite{nayar2003emor}.  The color of each curve is the accuracy of the shadow classification task after 
		 distorting the sequence by that response function.  The plots are split into three
		 separate plots using the same colormap for easier visualization.
		 In~\subref{subfig:datasize}, we simultaneously vary the number of images and length of
         time used in the sequence and report accuracy of the shadow-or-not classification task.
         }
         \end{center}
\end{figure*}

\subsection{Implementation Details}
We repeatedly alternate between the expectation and maximization step until the $S_t(\mathbf{x})$ labels do not change from
iteration to iteration, or until 50 iterations have passed.  In practice, most pixels' labels converge quickly.  In all 
experiments, more than 50\% of the pixels reach convergence before 6 iterations, and 99\% reach convergence before 20 iterations.

In practice, the linear system in Equation~\ref{eqn:leastsquares} can quickly become rank-deficient; for example, if one sets $S_t(\mathbf{x}) = 0$
for all $t$, the system over four variables reduces to rank one.  Intuitively, this makes sense, since recovery of surface normals
is numerically impossible for a pixel consistently under shadow.  When the assignment of $S_t(\mathbf{x})$ yields a singular matrix
in Equation~\ref{eqn:leastsquares}, we tried many methods of resetting $S_t(\mathbf{x})$
to regain full-rank, including a full reset $S_t(\mathbf{x}) = 1$ (i.e., pixel $\mathbf{x}$ is directly lit all the time) 
or $S_t(\mathbf{x}) = 0$ (pixel $\mathbf{x}$ always shaded, effectively giving up on estimating albedo and normals for this pixel).
However, we found that the most accurate results came from an 
incremental re-assignment,
which chooses the time $t$ so that the pixel $\mathbf{x}$ is brightest yet shadowed and reassigns $S_t(\mathbf{x}) = 1$, 
repeating until the resulting linear system is full rank.  This reassignment is done at each iteration before performing 
the expectation step\footnote{After a pixel has been reassigned and gone through one iteration of expectation and maximization 
steps, if its resulting labeling is equal to the labeling before reassignment, we declare that pixel as converged and accept
that labeling.  Therefore, a pixel under shadow at all times will be correctly labeled, albeit with a rank deficiency.}.

We additionally experimented with many initialization procedures, including assuming all pixels are always directly lit, and
random initialization.  The best method we found was to initialize $S_t(\mathbf{x}) = 1$, except for the $t$ where $I_t(\mathbf
{x})$ is minimal, using $S_t(\mathbf{x}) = 0$.

In real sequences, pixels become saturated as $I_t(\mathbf{x}) = 255$, breaking the color linearity assumption of 
Equation~\ref{eqn:lambertian_model}.   In this work, we replace any saturated $I_t(\mathbf{x})$ with the expected intensity consistent with the color linearity assumption; see the supplemental material for details.

Runtime is largely dependent on the number of non-sky pixels and the length of the image sequence.  To give a rough estimate
for runtime, performing this algorithm on 100 images at 512$\times$380 resolution (optimizing over 132,000 non-sky pixels) 
takes 2 minutes and 33 seconds on a 2.66GhZ Intel Xeon processor with 12 GB of memory across 8 cores.  Since the EM optimization
for each pixel is independent from any other pixel, each optimization is performed in parallel.

\section{Results}
\label{sec:results}
Here, we describe the results of our algorithm in controlled and uncontrolled environments, and show how this shadow estimation
procedures fits in with the larger field of outdoor photometric image formation.

\begin{figure*}[tp]
	\begin{center}
	\begin{tabular}{c|cccc}
	\begin{sideways} ~~~~~~~Example \end{sideways} & 
		\includegraphics[height=0.123\textheight]{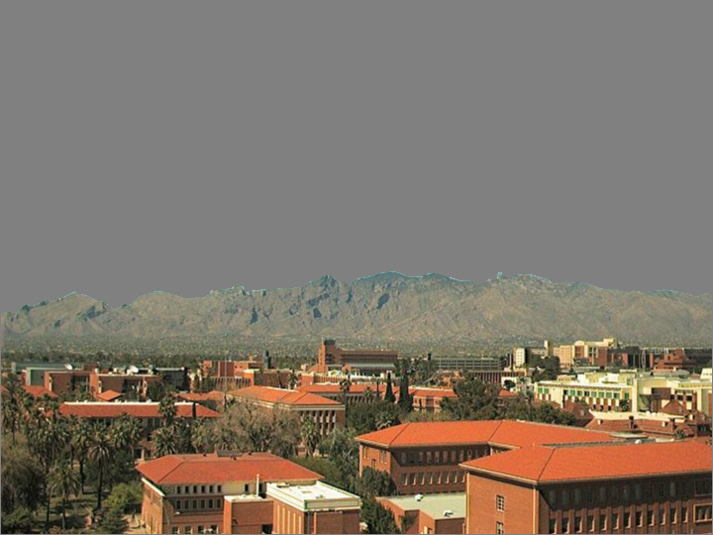} &
		\includegraphics[height=0.123\textheight]{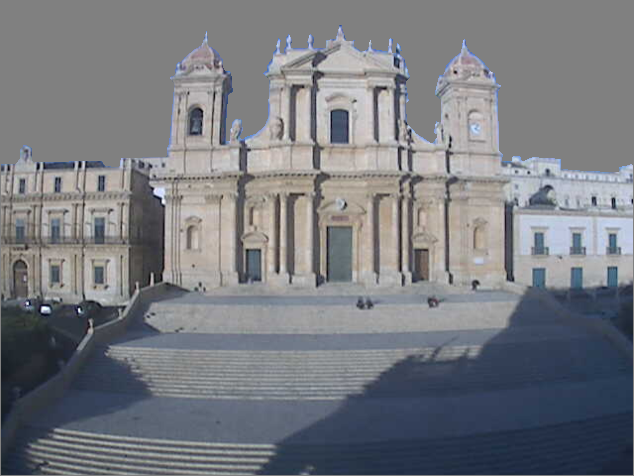} &
		\includegraphics[height=0.123\textheight]{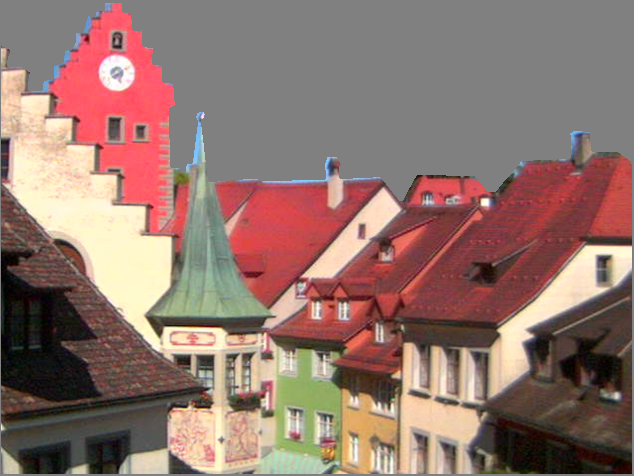} &
		\includegraphics[height=0.123\textheight]{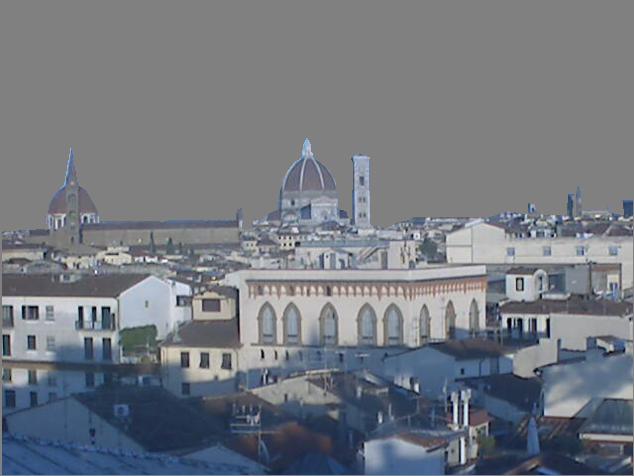} \\
	\hline
	\begin{sideways} ~~~~~~~~FTLV \cite{sunkavalli2007timelapse} \end{sideways}& 
		\includegraphics[height=0.125\textheight]{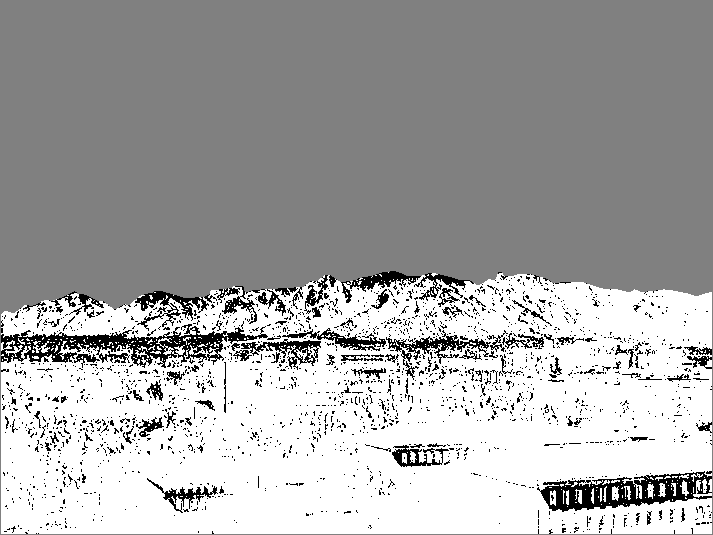} &
		\includegraphics[height=0.125\textheight]{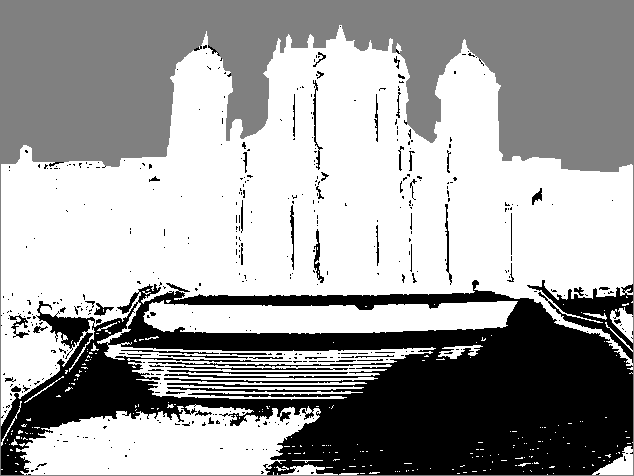} &
		\includegraphics[height=0.125\textheight]{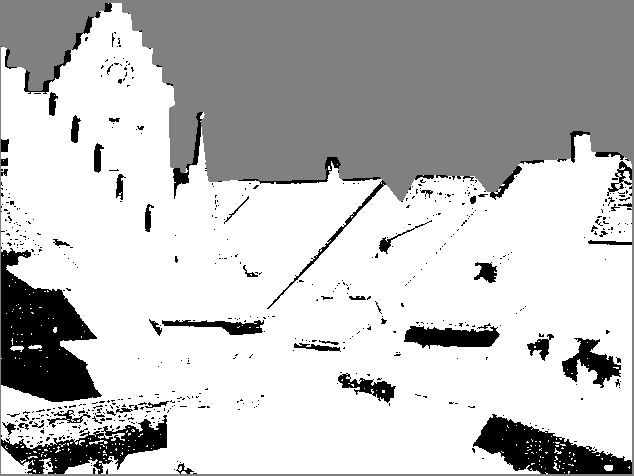} &
		\includegraphics[height=0.125\textheight]{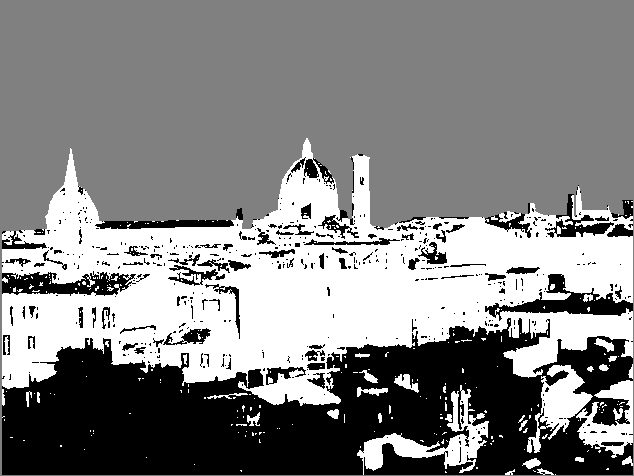} \\
	\hline
	\begin{sideways} ~~~~~~~~~HS \cite{abrams2012helio} \end{sideways}& 
		\includegraphics[height=0.125\textheight]{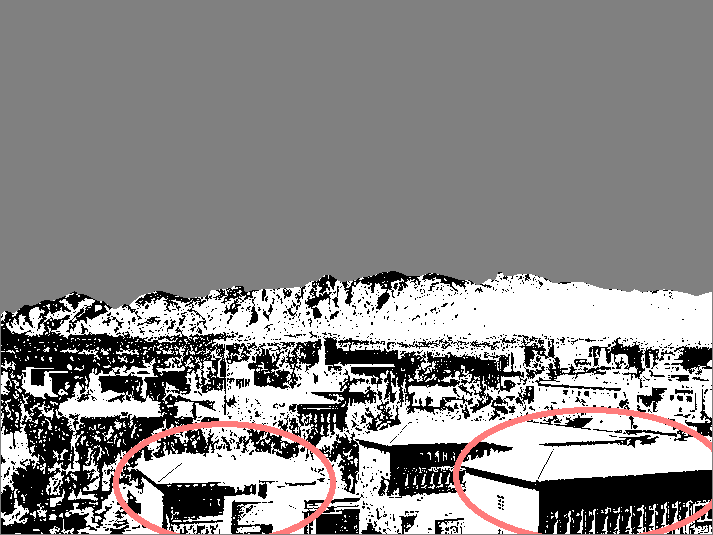} &
		\includegraphics[height=0.125\textheight]{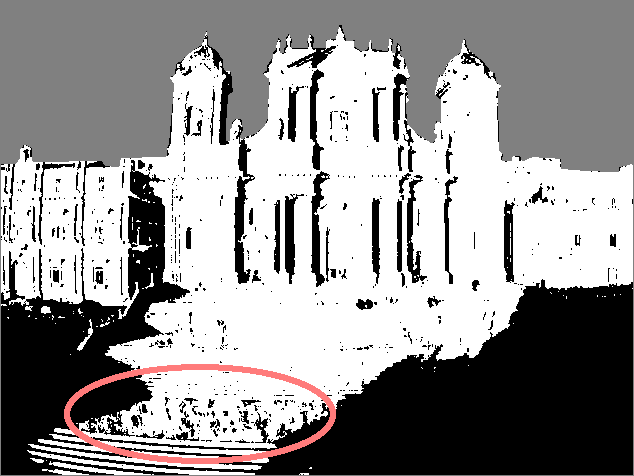} &
		\includegraphics[height=0.125\textheight]{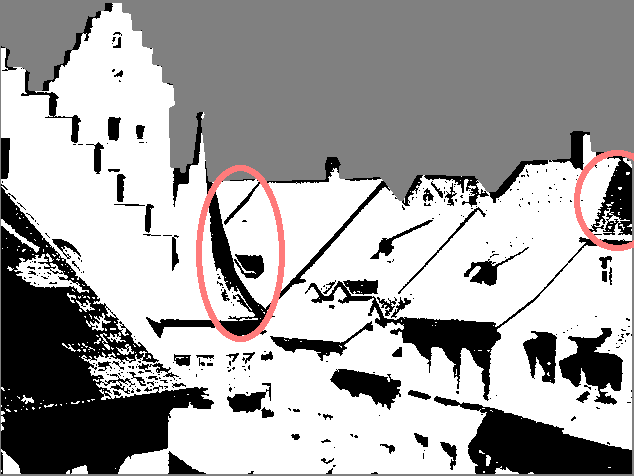} &
		\includegraphics[height=0.125\textheight]{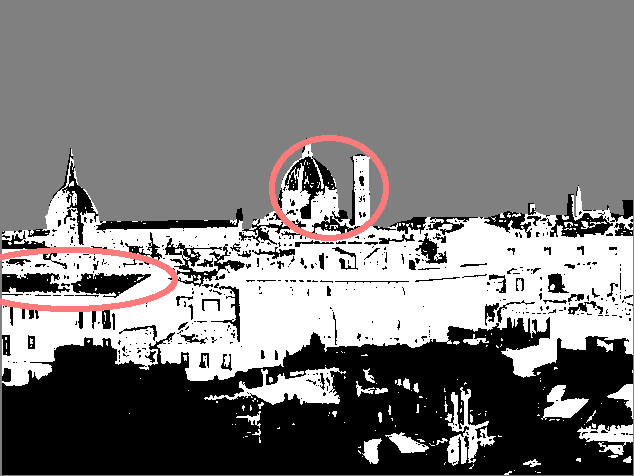} \\
	\hline
	\begin{sideways} ~~~~~~~Our Result \end{sideways} & 
		\includegraphics[height=0.125\textheight]{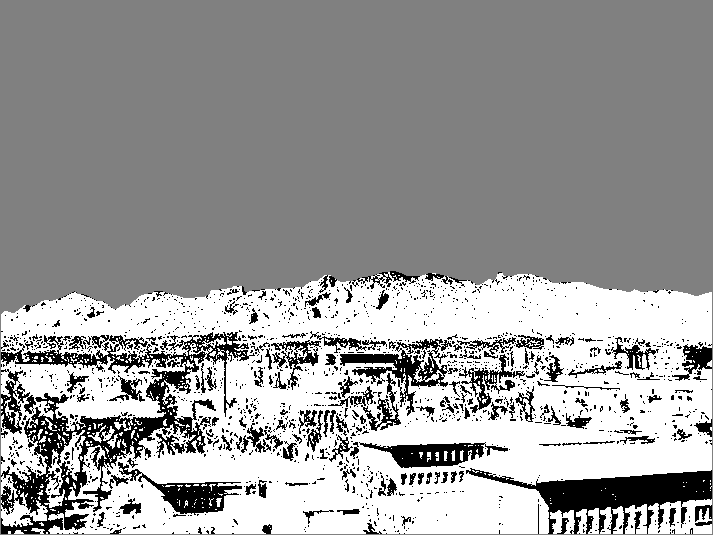} &
		\includegraphics[height=0.125\textheight]{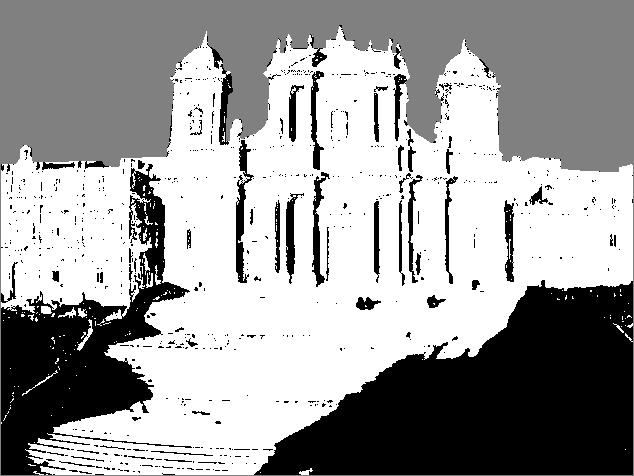} &
		\includegraphics[height=0.125\textheight]{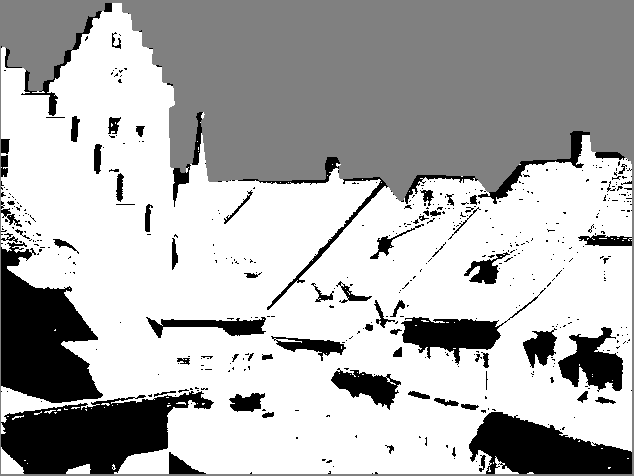} &
		\includegraphics[height=0.125\textheight]{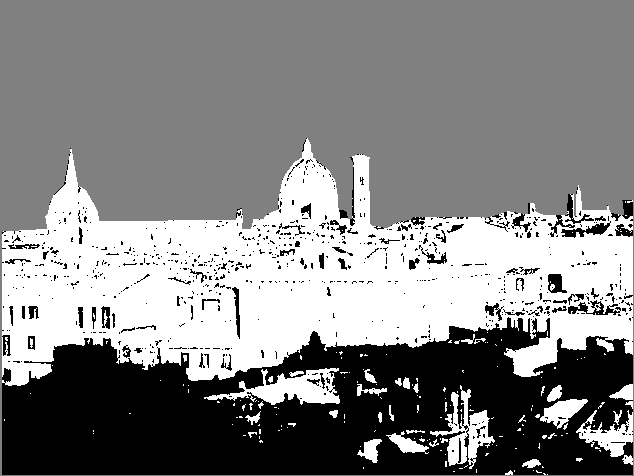} \\
	\begin{sideways} ~~~~~~~~Normals \end{sideways} &
		\includegraphics[height=0.125\textheight]{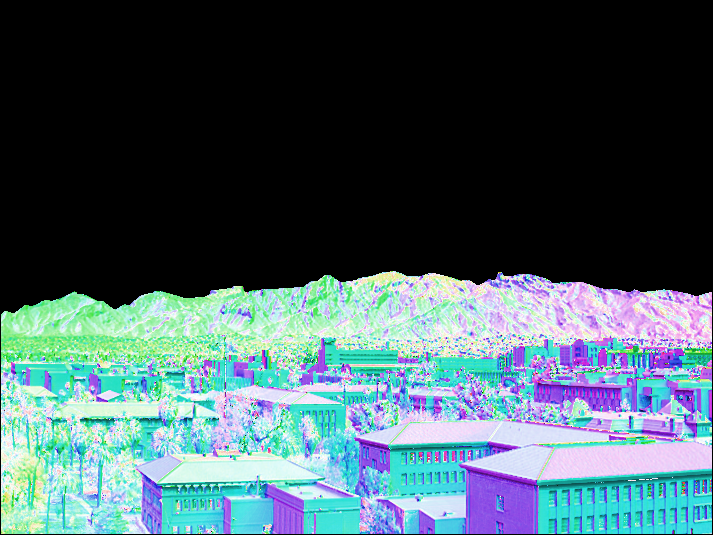} &
		\includegraphics[height=0.125\textheight]{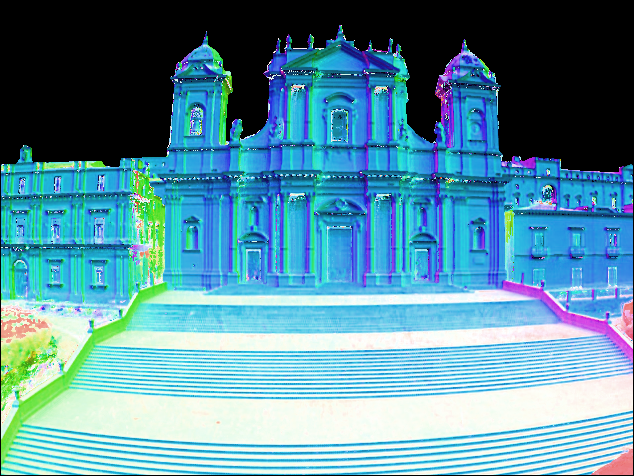} &
		\includegraphics[height=0.125\textheight]{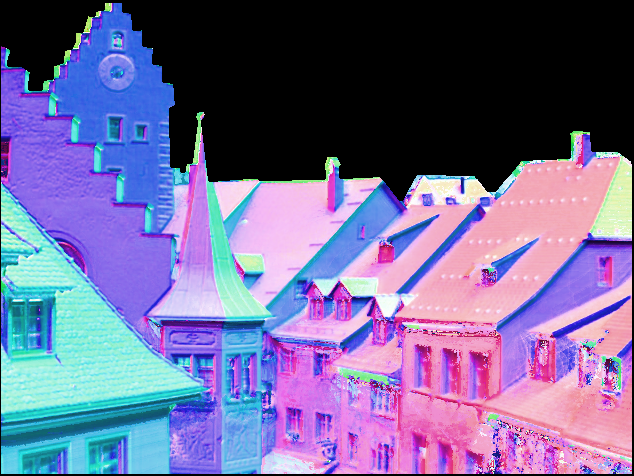} &
		\includegraphics[height=0.125\textheight]{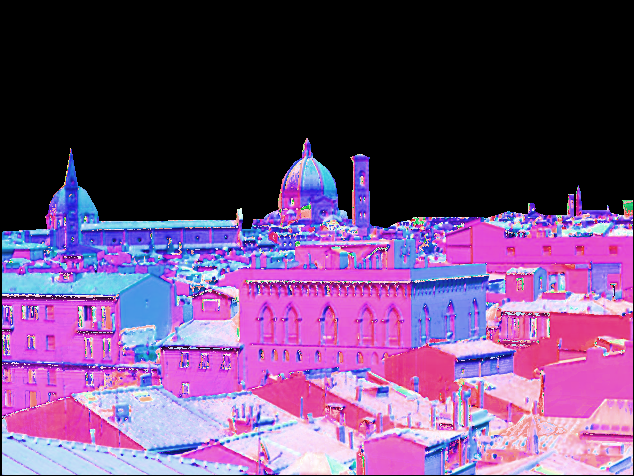} \\
	\begin{sideways} ~~~~~~~~~Albedo \end{sideways} &
		\includegraphics[height=0.125\textheight]{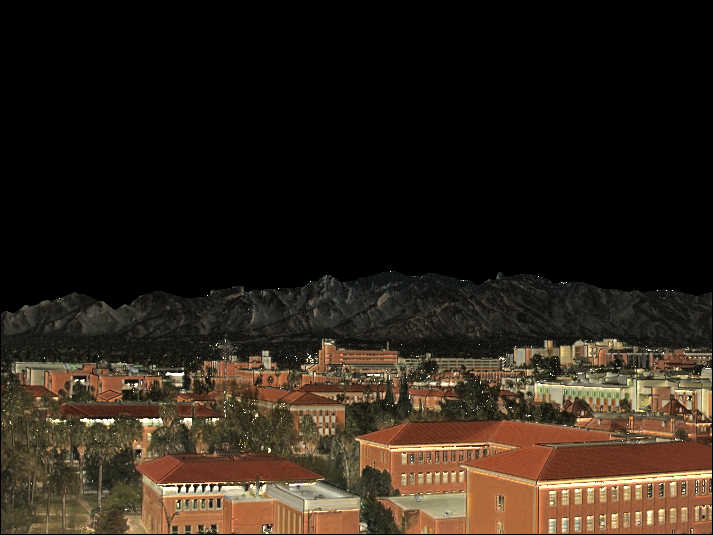} &
		\includegraphics[height=0.125\textheight]{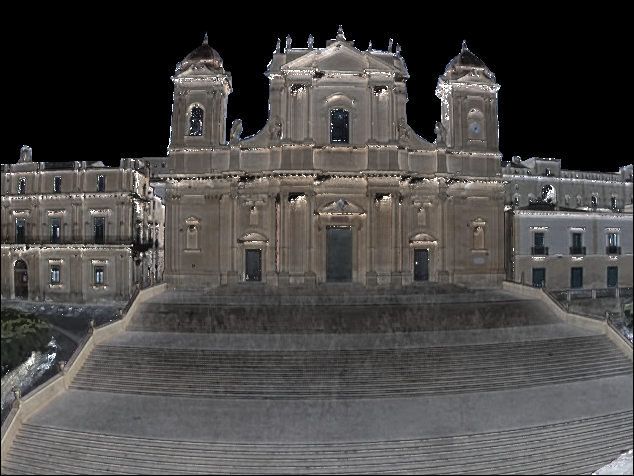} &
		\includegraphics[height=0.125\textheight]{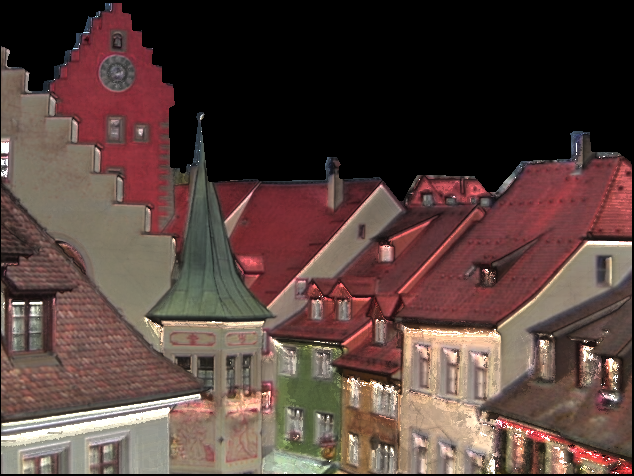} &
		\includegraphics[height=0.125\textheight]{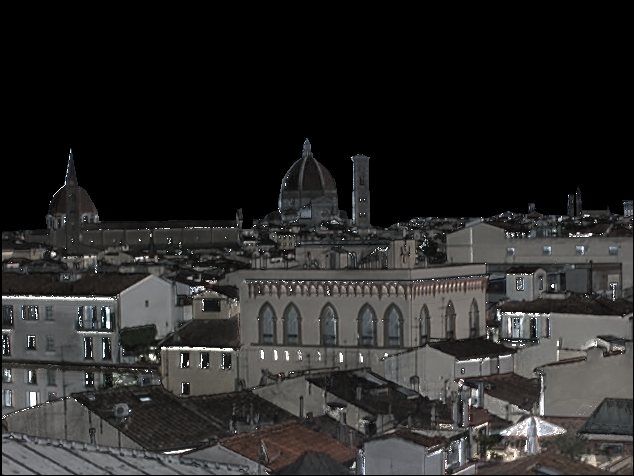} \\
	\end{tabular}
	\vspace{1.5em}
	\caption{Results from various shadow estimation approaches, with imagery taken from the AMOS dataset~\cite{jacobs07amos}.
			 From top to bottom: an original image, the Factored Time Lapse Video approach~\cite{sunkavalli2007timelapse}, the 
			 Heliometric Stereo approach~\cite{abrams2012helio}, and our results.  Each shadow estimation approach shows the
			 estimated shadow mask for the given original frame.  Although the approach from~\cite{abrams2012helio} does better, it makes many subtle 
			 errors (circled in pink).  Our approach makes fewer mistakes than the other 
			 two methods,
			 and accurately recovers both large-scale and fine detail in shadow patterns. Although our goal is to recover 	
			 shadows, our EM algorithm simultaneously recovers normals and albedos as a byproduct (last two rows).  We show
			 normals in an East-North-Up coordinate frame, using
			 the color map from~\cite{abrams2012helio}, where the hue codes for geographic orientation and the lightness
			 represents degrees away from the zenith.}
			 	\label{fig:fullpage}

	\end{center}
\end{figure*}

\subsection{Controlled Environments}
To test our method, we created a synthetic dataset using the image formation model of Section~\ref{sec:methods}, simulating
lighting directions from a virtual camera\footnote{We originally considered using the
available ground truth synthetic data from~\cite{isaza2012synthetic}, but their ground truth labels only include cast shadows, 
never attached shadows, and our algorithm does not make such a distinction.}.  Using a rendering pipeline to create 300 images over the 
span of a simulated year, we ran our algorithm and recovered the exact solution that generated the data; see 
Figure~\ref{fig:synthetic_reconstruct}.  We recover the correct shadows
with 99.79\% accuracy, with most pixels never making a single mistake.  Although the goal is to recover shadows,
we also recover the correct albedo and normals to 0.29 intensities (from 0-255) and 0.20$^\circ$, respectively.

Real webcam images usually suffer from nonlinear camera response, but our image formation model does not account for such distortions.
Including unknown response in our model would make the optimization more complex, as a small change in the response affects all
pixels in the timeseries.
To test robustness against the unknown camera response, we distorted our data by all of the functions from the Database of Response
Functions from Grossberg and Nayar~\cite{nayar2003emor} and re-ran our shadow estimation algorithm.  As shown in Figure~\ref
{subfig:responsefns}, nonlinear camera response has a negligible effect on shadow estimation: applying a response function to synthetic
data usually decreases accuracy by less than 1\%, and at worst, less than 5\%.

As noted in~\cite{abrams2012helio}, accurate recovery of surface normals from a limited data set is challenging.  Therefore,
we perform experiments to see how long or short of a time period is required to recover accurate shadow volumes.  We take
a sequence and perform the proposed approach on different numbers of images $n$, as well as the lengths of 
the original sequence  (e.g. an hour, a day, a week).  We repeat the experiment for different random subsets of imagery and report 
average accuracy over 10 trials.  Figure~\ref{subfig:datasize}
demonstrates that while having a longer clock time improves the result, the number of images used is much more important,
as long as the original sequence has at least 25 images across a few hours, we reliably recover the correct shadow volume.
For all results presented in this paper, we use 50-100 images taken over the span of many months.

These results contradict~\cite{abrams2012helio}, which states that
explicit radiometric modeling and input imagery spanning many weeks is necessary to recover good surface normals. We attribute this 
property to the difference in appearance for shaded vs. directly-lit pixels:
although we may not know what surface normal or radiometric curve describes the intensity of a pixel under direct sunlight, the
intensity difference between a pixel's expected intensity in and out of shadow is substantial enough to accurately disambiguate the two (given a 
lighting direction).

\subsection{Uncontrolled Environment}

The experiments in the previous section describe how our algorithm performs in synthetic environments.
The real world, however, has many error modes that cannot easily be exhaustively enumerated in such an environment.
Given that our goal is to recover shadows from real time-lapse sequences, we report performance of shadow
classification with respect to real scenes, taken from the Archive of Many Outdoor Scenes (AMOS) webcam dataset~\cite{jacobs07amos}.

\subsubsection{The Labeled Shadows in the Wild Dataset}
To report quantitative measurements, we selected 7 scenes from the AMOS dataset and labeled ground truth shadow masks for 50 images 
each.  Our labeled data comes
from a variety of cameras across the globe, including a busy plaza in Barcelona, a university
in Arizona, and a camera in Germany that observes complicated geometry.  These cameras break many of the
assumptions that we make in our shadow estimation procedure, through the inclusion of atmospheric haze,
time-variable geometry (most notably in the plaza where pedestrians and flea markets occupy large parts of the image),
variable exposure, and nonlinear camera response.  

Labeling such a dataset is itself nontrivial.  In selecting
images from each camera, we use the automatic image selection algorithm of~\cite{ackermann2012ps} to avoid human
bias and select clear-day images with easily-discriminable shadow boundaries.  The scenes we use demonstrate considerable
complexity in scene geometry and noise factors, and hand-labeling millions of pixels is challenging for even the
most experienced graduate student.  We therefore take advantage of the natural color distribution of shadows described
in~\cite{hoeim2011shadows} and label only a sparse sample of pixels for which we are confident.  These sparse labels
are then propagated using the matting equation of~\cite{levin2006matting}, creating an $\alpha$-mat
across the image.  We label any pixel with $\alpha > 0.7$ as directly lit, and $\alpha < 0.3$ as shadowed.  All
other pixels are left unlabeled.  All labels were hand-verified before experimentation.  See the
supplemental material for example labels.

To facilitate future comparison in time-lapse shadow estimation, the Labeled Shadows in the Wild (LSW) dataset and our code is available
at [anonymized].

\subsubsection{Evaluation}
To give the best possible performance of alternative existing algorithms, we use the LSW data for cross-validation and then test on
the same dataset.  We choose parameters
with three strategies.  The ``Suggested'' strategy uses the parameters prescribed in their respective papers.  For the ``Global'' strategy, we select a single set of parameters that maximizes each algorithm's average
accuracy across the 7 scenes.  For the ``Optimal'' strategy, we tweak parameters for each scene individually to maximize accuracy.
As our approach is parameter-free, no cross-validation is necessary.

Interestingly, the parameters suggested by~\cite{abrams2012helio, sunkavalli2007timelapse}, originally set empirically, are not 
quite the same as
recovered in the cross-validation step.  Although the FTLV approach has suggested parameters $(\theta_p, \theta_k) = (0.2,1.5)$,
we found that for this dataset, the best parameters were $(0.05, 1.5)$.  The HS approach
suggests $(\theta_p, \theta_\lambda) = (0.8, 0.05)$, whereas the optimal values we use are $(0.98, 0.01)$.

Our quantitative results are shown in Table~\ref{table:quant}.  These show that our
approach and \cite{abrams2012helio} perform roughly equally when parameters have been chosen to maximize accuracy
per dataset.  However, we stress that these numbers represent the best-case
performance of the other methods, and that using any other parameter setting deteriorates their performance,
sometimes dramatically (in the case of the default suggested parameters).

\begin{table}[t]
\begin{center}

\begin{tabular}{|c|c|c|c|}
\hline
Algorithm & Suggested & Global & Optimal \\
\hline
FTLV~\cite{sunkavalli2007timelapse} &  74.22 & 74.94 & 82.03 \\
HS~\cite{abrams2012helio} & 84.58 & 86.76 & \textbf{87.40} \\
Our approach & \textbf{87.13} & \textbf{87.13} & 87.13 \\
\hline
\end{tabular}

\caption{Accuracy of various approaches (higher is better) on the Labeled Shadows in the Wild dataset, in percent.
	     In the ``Suggested'' column, we use the parameters as reported in their respective papers.  
	     In ``Global'', we set the parameters of the first two approaches by treating the test dataset as a 
	     cross-validation set to maximize accuracy across all scenes.  Finally, in ``Optimal'', we optimize
	     a set of parameters separately for each scene. Our approach is parameter-free and does not require any such
	     cross-validation.}
\label{table:quant}
\end{center}
\vspace{-1em}
\end{table}

A qualitative comparison is shown in Figure~\ref{fig:fullpage}.  We select 100 images from several cameras from the AMOS
dataset~\cite{jacobs07amos}, again using the image selection algorithm of~\cite{ackermann2012ps}, a multi-scale alignment
procedure from~\cite{jacobs09adventures} and perform each of shadow estimation procedures.  Because the FTLV algorithm is 
designed for a single day's worth of imagery, the resulting shadow masks are understandably unusable.  The HS approach 
does better, but often shades too much of the scene (most visibly in columns 1 
and 3 of Figure~\ref{fig:fullpage}).

\subsection{Initialization for Time-Lapse Photometric Analysis}

\begin{figure*}[t]
\begin{center}
\subfigure[Initialized using~\cite{abrams2012helio}]{
			\shortstack{\includegraphics[width=0.18\textwidth]{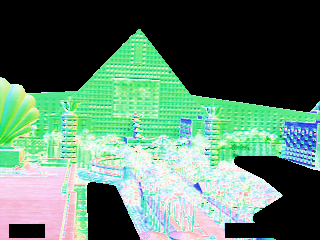}}
			\label{subfig:oldinit}}
\subfigure[Initialized using our approach]{
			\shortstack{\includegraphics[width=0.18\textwidth]{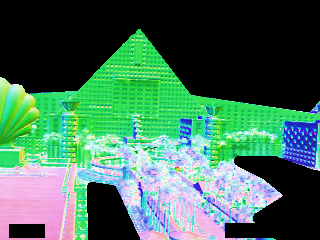}}
			\label{subfig:newinit}}
\subfigure[Ground truth]{
			\shortstack{\includegraphics[width=0.18\textwidth]{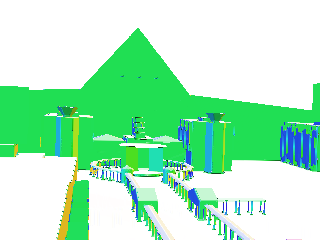}}
			\label{subfig:ground_truth_normals}}
\subfigure[Errors using ~\cite{abrams2012helio}] {
			\shortstack{\includegraphics[width=0.18\textwidth]{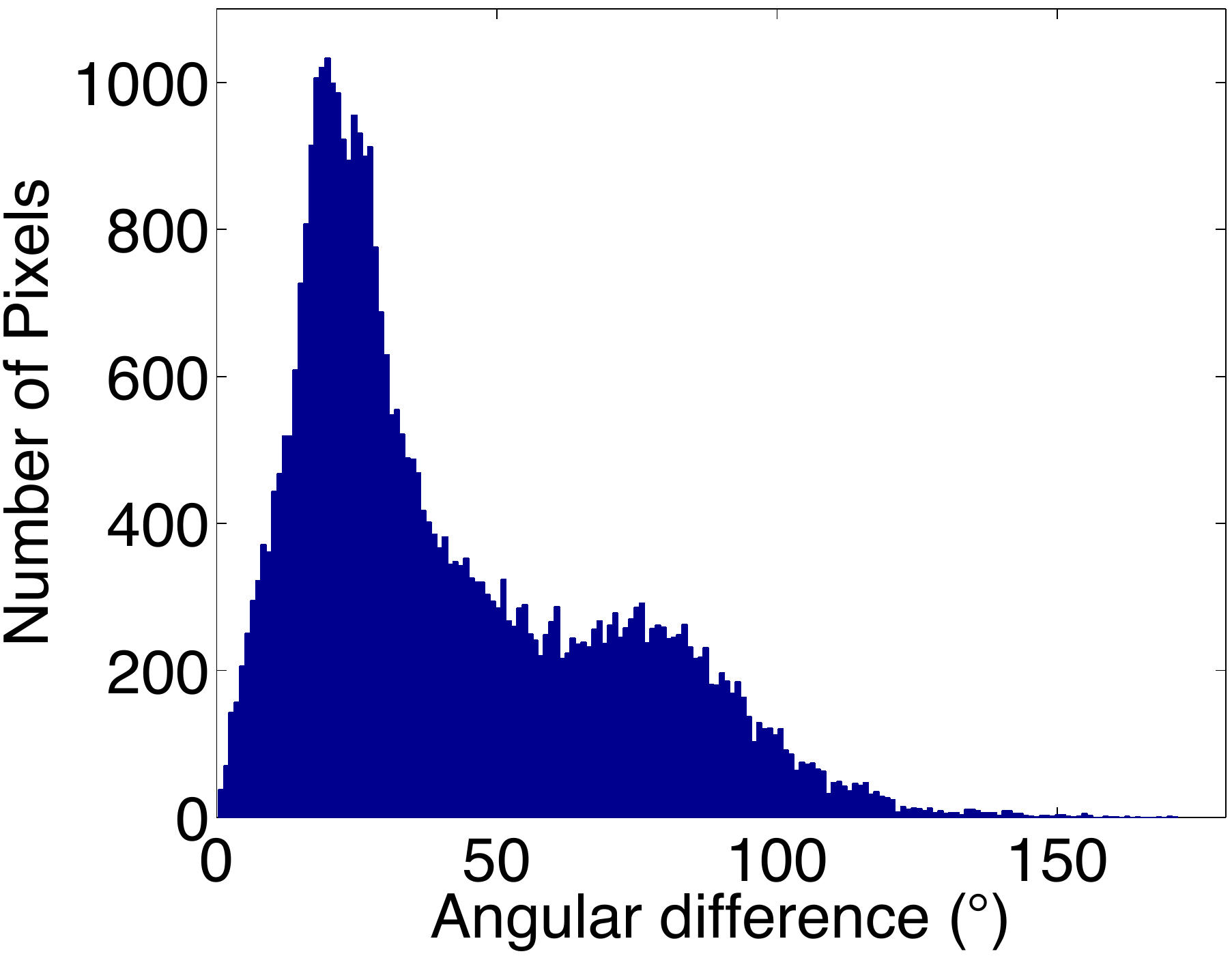}}
			\label{subfig:olderror}}
\subfigure[Errors using our approach] {
			\shortstack{\includegraphics[width=0.18\textwidth]{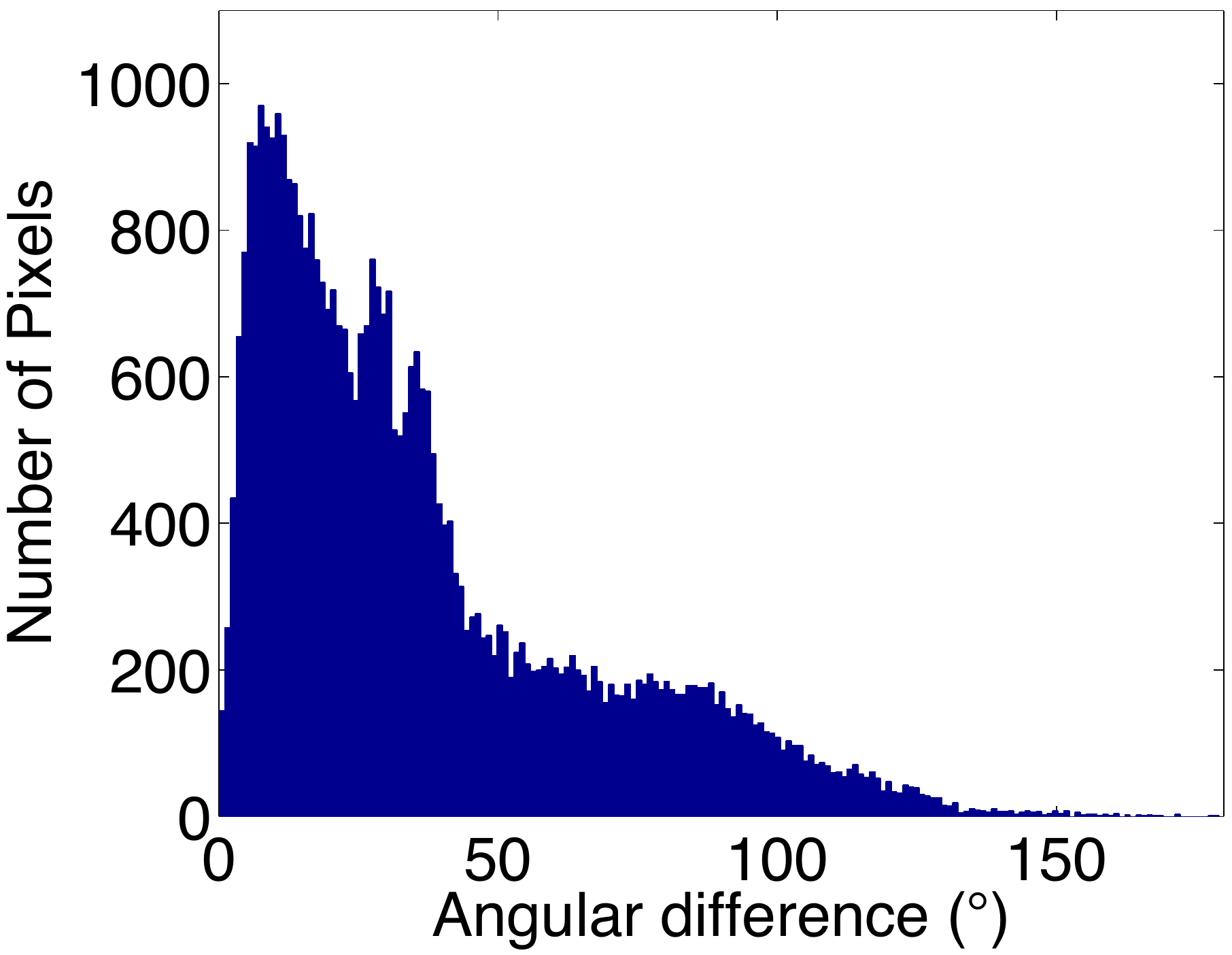}}
			\label{subfig:newerror}}
\caption{Comparing errors in normal estimates by initializing the optimization in~\cite{abrams2012helio}
		 with their suggested method~\subref{subfig:oldinit} and the proposed method~\subref{subfig:newinit}.  
		 While ground truth data from Google Earth~\subref{subfig:ground_truth_normals} does not contain real-world normal 
		 variation in  such as tree leaves and individual windows, it provides a convenient surrogate for estimating normal 
		 accuracy for scenes in the wild. Although both approaches appear to have substantial errors, as noted 
		 in~\cite{abrams2012helio}, these largely come from low-detail polygons in Google Earth models.  Initializing the 
		 optimization with our approach, however, substantially decreases angular error~\subref{subfig:olderror} 
		 \subref{subfig:newerror}.
		   }
\label{fig:normal_accuracy}
\end{center}
\end{figure*}

Estimating the shadow volume for a sequence is often a first step for more in-depth photometric analysis of
a time-lapse scene~\cite{abrams2012helio, ackermann2012ps,sunkavalli2007timelapse}.  In each of these works,
shadow estimation is considered a pre-process during the initialization step.
We use the code from~\cite{abrams2012helio}, which solves for surface normals and albedo from a time-lapse
scene, but also simultaneously recovers estimates for per-image exposure and radiometric response functions.
To test the practicality of our routine, we initialize the optimization in two different ways: one using their
suggested initialization, and another using our proposed approach.  We then let the
optimization continue until convergence; the only difference is in
the initialization routine.

We compare the resulting normals from each optimization
to Google Earth ground truth models.
This comparison, visualized in Figure~\ref{fig:normal_accuracy}, shows a few important details.  First,
although the errors for both initialization routines appear very large,
most of these errors are coming from areas of the scene not modeled well by Google Earth,
and the errors from well-modeled surfaces are less than 10 degrees.  Further, initializing with
the proposed EM algorithm yields much more accurate surface normals than previously reported: the
peak of surface normal error shifts from 20 degrees to less than 10 degrees. This emphasizes
the importance of estimating shadows in these larger pipelines.  

\section{Conclusions}

In this work, we introduce a method for classifying a pixel as being directly lit or in shadow in real outdoor time-lapse
sequences.  Our expectation-maximization approach is parameter-free and outperforms previous methods.  
To validate our algorithm, we perform synthetic experiments to show that our
approach is robust to nonlinear camera response and is invariant to sequence length.  We also introduce the Labeled Shadows in 
the Wild dataset, which offers a standard basis for future work to evaluate shadow estimation in the face of noisy signals in real
outdoor scenes.  We show that our approach improves normal field
accuracy when used as an initialization step for richer image formation model inference.

Detecting shadows is a critical piece of any visual system, and although previous state-of-the-art clever thresholding works well in
some circumstances, optimizing over the shadow process in the image formation
model is an important part of outdoor time-lapse analysis.

{\small
\bibliographystyle{ieee}
\bibliography{visionRefs}
}

\pagebreak
\appendix

{\Large Supplemental Material}

\section{Saturated Pixels}

As described in the main body of the text, real cameras tend to have saturated color channels where 
$I_t(\mathbf{x}) = 255$, breaking color linearity.  Note that we can 
rewrite the image formation model as
\[
I_t(\mathbf{x}) = c(\mathbf{x})\alpha_{xt}
\]
for some color vector $c$ and scalar $\alpha$.  Therefore, if the images fit our model, the unsaturated
colors are linear in RGB space.

To handle saturation, we replace any saturated $I_t(\mathbf{x})$ with the intensity that best fits the
color linearity model.  We first estimate the color vector $c(\mathbf{x})$ of each pixel as the mean of $I(\mathbf{x})$, 
excluding any $t$s where any channel of $I_t(\mathbf{x})$ is saturated.  Then, we estimate the per-time scalar $\alpha_t$
as the solution to the linear system $I_t(\mathbf{x}) = \alpha c(\mathbf{x})$, again excluding any saturated channels.
Finally, we replace any saturated $I_t(\mathbf{x})$ as $\alpha c(\mathbf{x})$.  If each of the color channels are saturated
(i.e., the pixel is pure white), then we fix that pixel as being directly-lit, and do not attempt
to optimize its label.

\section{Labeled Shadows in the Wild}

Here, we provide example labels from the Labeled Shadows in the Wild dataset,
described in the main body of the text.  Each figure contains two images from a single camera, and
shows an example image and its label.  On the example image, violet borders indicate
unknown pixel intensities, due to timestamps and alignment.  On the labeled image, black indicates
a shadowed pixel, white indicates a pixel under direct illumination from the sun, and gray values are unknown.

\begin{figure*}
\begin{center}
\includegraphics[height=0.145\textheight]{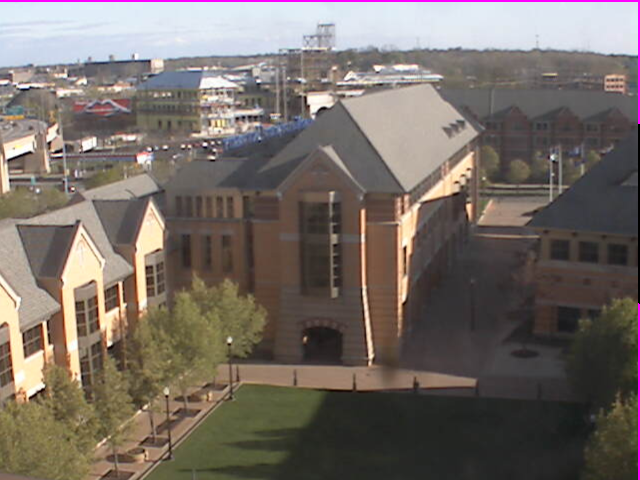}~~\includegraphics[height=0.145\textheight]{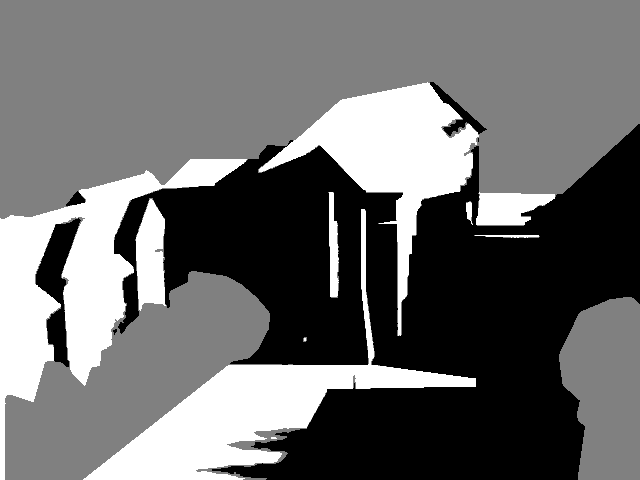}~~
\includegraphics[height=0.145\textheight]{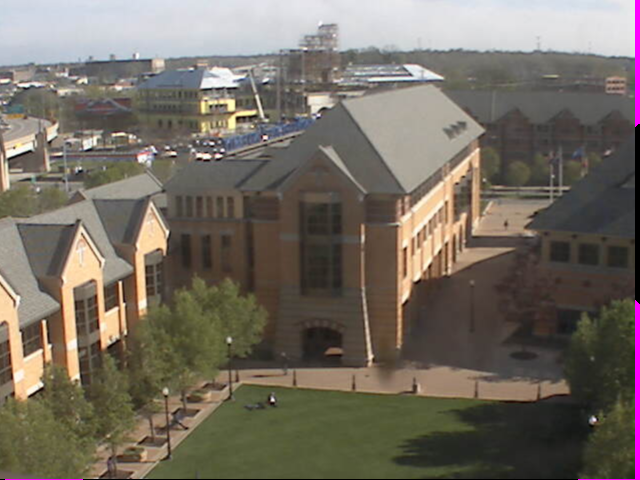}~~\includegraphics[height=0.145\textheight]{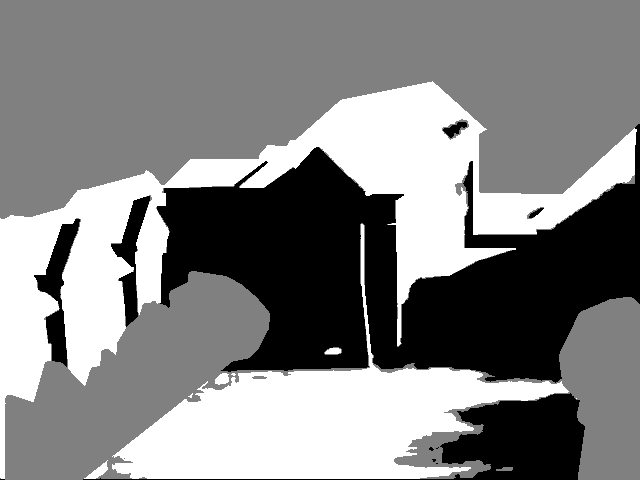}
\caption{Labeled results from a webcam in Columbia, Missouri.}
\label{fig:1092}
\end{center}
\end{figure*}

\begin{figure*}[t]
\begin{center}
\includegraphics[height=0.145\textheight]{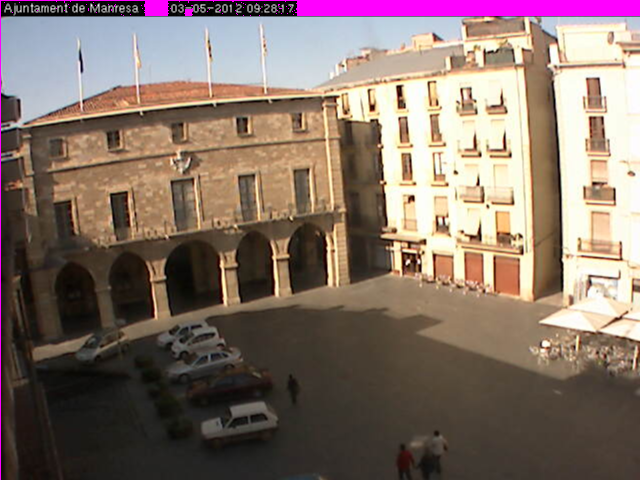}~~\includegraphics[height=0.145\textheight]{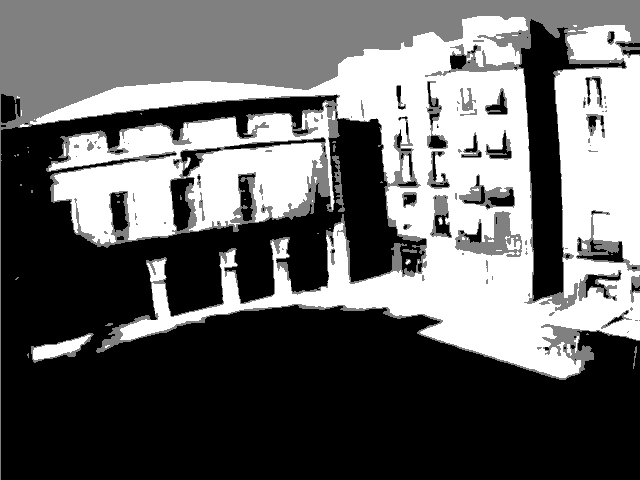}~~
\includegraphics[height=0.145\textheight]{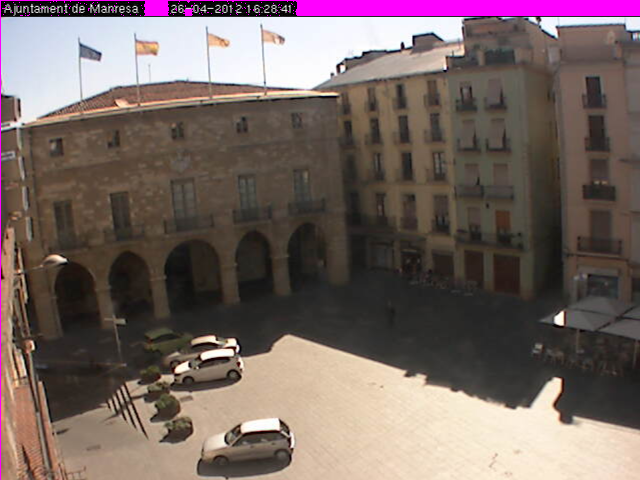}~~\includegraphics[height=0.145\textheight]{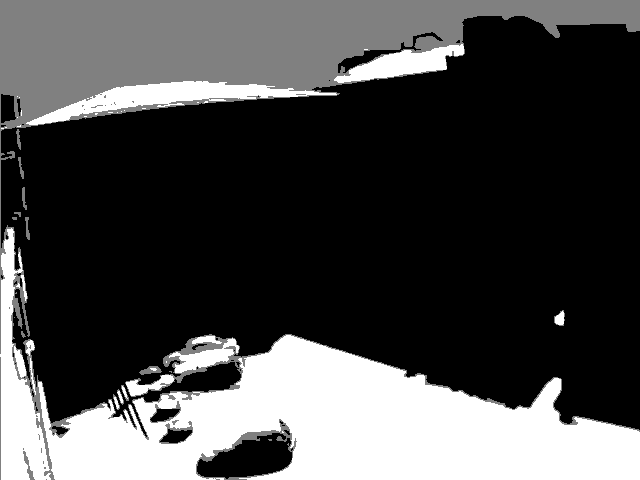}
\caption{Labeled results from a webcam at a plaza in Barcelona.}
\label{fig:1290}
\end{center}
\end{figure*}

\begin{figure*}[t]
\begin{center}
\includegraphics[height=0.145\textheight]{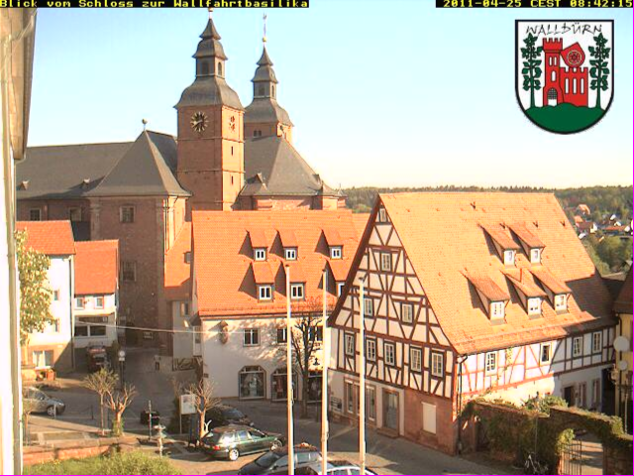}~~\includegraphics[height=0.145\textheight]{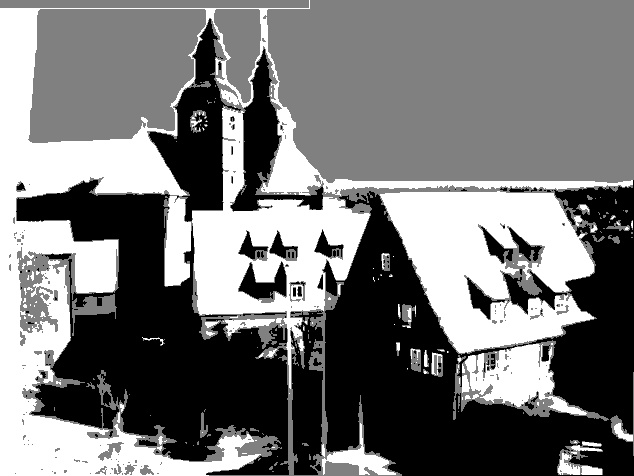}~~
\includegraphics[height=0.145\textheight]{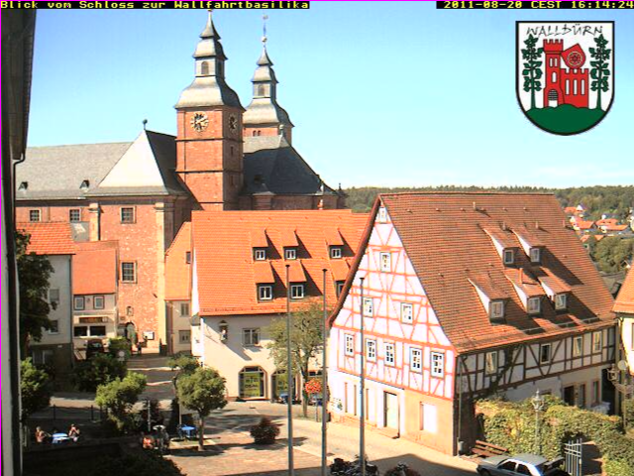}~~\includegraphics[height=0.145\textheight]{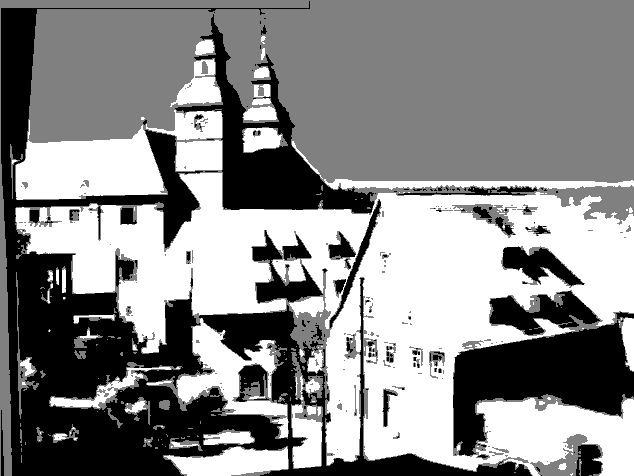}
\caption{Labeled results from a webcam in Walld\"{u}m, Germany.}
\label{fig:4679}
\end{center}
\end{figure*}

\begin{figure*}[t]
\begin{center}
\includegraphics[height=0.145\textheight]{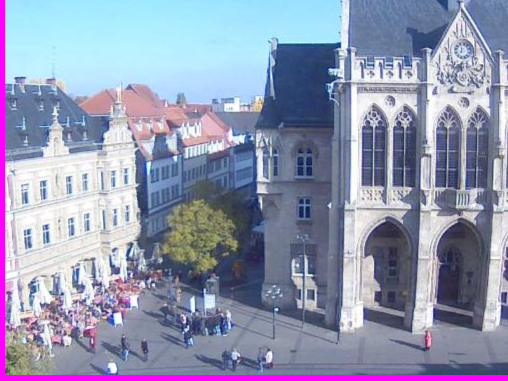}~~\includegraphics[height=0.145\textheight]{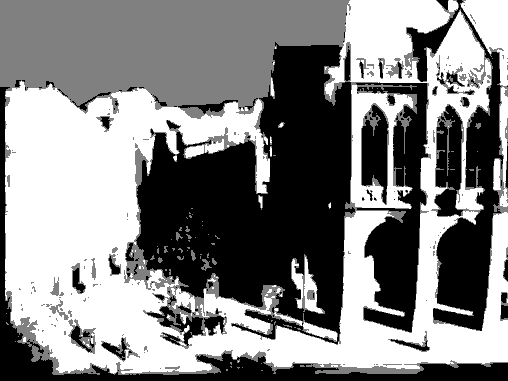}~~
\includegraphics[height=0.145\textheight]{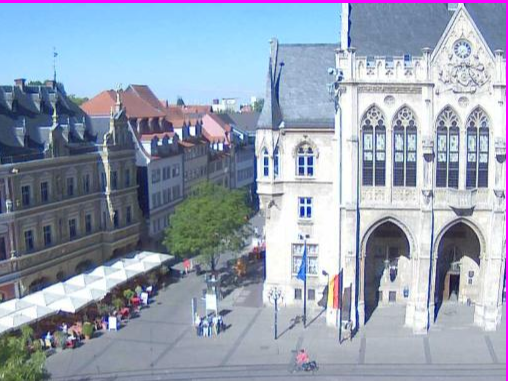}~~\includegraphics[height=0.145\textheight]{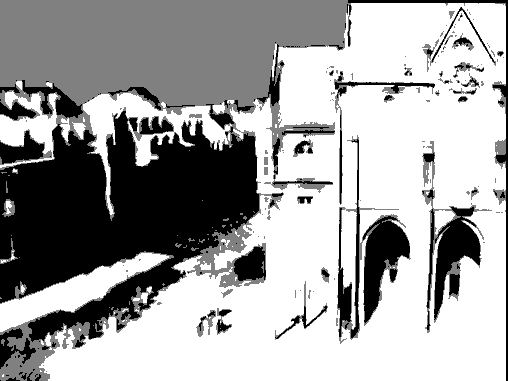}
\caption{Labeled results from a webcam in Erfurt, Germany.}
\label{fig:4795}
\end{center}
\end{figure*}

\begin{figure*}[t]
\begin{center}
\includegraphics[height=0.145\textheight]{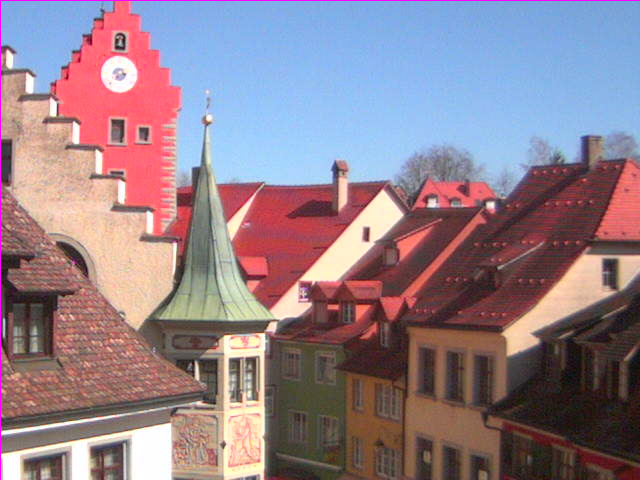}~~\includegraphics[height=0.145\textheight]{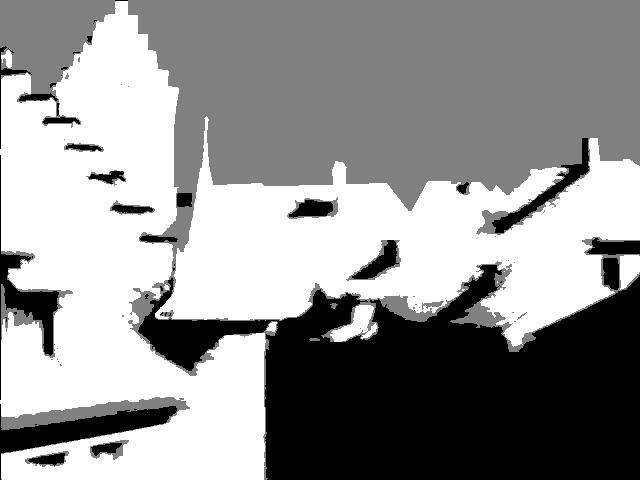}~~
\includegraphics[height=0.145\textheight]{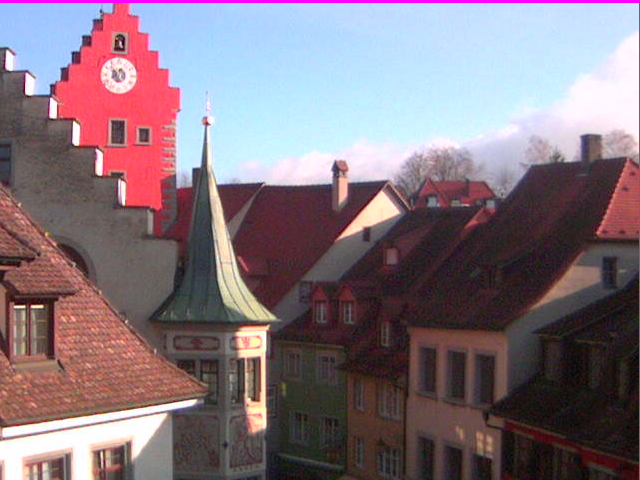}~~\includegraphics[height=0.145\textheight]{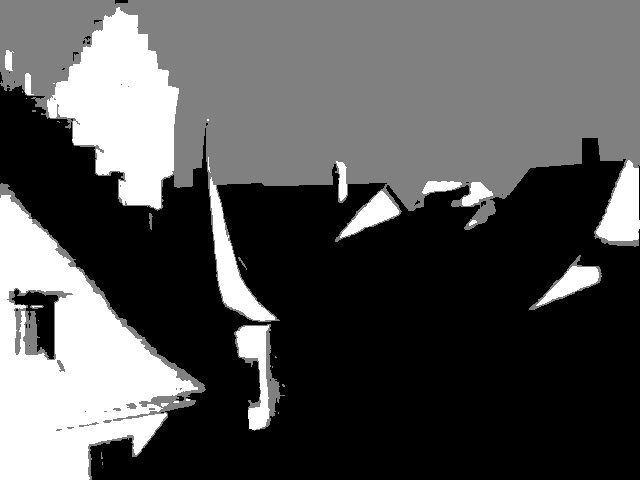}
\caption{Labeled results from a webcam in Meersburg, Germany.}
\label{fig:10870}
\end{center}
\end{figure*}

\end{document}